\documentclass[conference, 10pt]{IEEEtran}

\IEEEoverridecommandlockouts                              

\usepackage[square,numbers]{natbib}
\usepackage{times}
\usepackage{amssymb}

\usepackage{graphics} 
\usepackage{epsfig} 
\usepackage{amsmath} 
\usepackage{multirow}
\usepackage{tabularx}
\usepackage{booktabs}
\usepackage{tabulary}
\usepackage{graphicx}
\usepackage{footnote}
\usepackage[table]{xcolor}

\usepackage[]{algorithm2e}

\usepackage[font=small]{caption}
\DeclareCaptionFont{tiny}{\tiny}

\title{\LARGE \bf
Joint Pose and Principal Curvature Refinement Using Quadrics
}

\author{Andrew Spek$^{1}$ and Tom Drummond$^{2}$%
	\thanks{$^{1}$Andrew Spek is with Faculty of Electrical Engineering, Monash University
        {\tt\small andrew.spek@monash.edu}}%
    \thanks{$^{2}$Tom Drummond is with Faculty of Electrical Engineering, Monash University
        {\tt\small tom.drummond@monash.edu}}%
}

\begin{document}

\maketitle 
\thispagestyle{empty}
\pagestyle{empty}


\begin{abstract}
	In this paper we present a novel joint approach for optimising surface curvature and pose alignment. We present two implementations of this joint optimisation strategy, including a fast implementation that uses two frames and an offline multi-frame approach. We demonstrate an order of magnitude improvement in simulation over state of the art dense relative point-to-plane Iterative Closest Point (ICP) pose alignment using our dense joint frame-to-frame approach and show comparable pose drift to dense point-to-plane ICP bundle adjustment using low-cost depth sensors. Additionally our improved joint quadric based approach can be used to more accurately estimate surface curvature on noisy point clouds than previous approaches.
\end{abstract}



\section{Introduction}

Surface curvature is a geometric feature that has been shown to be useful for modelling \cite{Guillaume2004}\cite{Rusinkiewicza}\cite{Mitra2004}, robotic control \cite{Rusu2010a}\cite{Sanchez-Fibla2013} and segmentation \cite{Lee2014}. Surface curvature is a measure of the rate at which the surface normal angle changes while travelling along the surface. This makes curvature a second order derivative, and therefore highly sensitive to noise in the surface. The principal curvatures represent the direction of maximal and minimal curvature at any point. The introduction of low-cost depth sensors, such as the Microsoft Kinect, provided an easy way to collect real-time depth information cheaply. However, the noise present in this sensor \cite{Khoshelham2012} has made it challenging to compute accurate curvature values directly \cite{Spek2015} due to its high noise sensitivity. 

Using quadrics has been shown to improve the robustness of curvature computations \cite{Besl1988}\cite{Douros2002} and advances in GPGPU programming has allowed for real-time dense estimation of curvature \cite{Griffin2012}\cite{Spek2015}. Using a quadric representation allows the principal curvatures to be extracted directly, through a simple manipulation of quadric parameters described in Section \ref{subsec:quad_surface}. Additionally constraints can be placed on the quadric representation to reduce the possible representable surfaces to improve the reliability of curvature estimates \cite{Spek2015}. A limitation of these methods is that they are designed for use on single-frame of live data or pre-registered low noise dense point clouds. Error in the registration accuracy or noise in the model can result in poor curvature estimates.

The Iterative Closest Point (ICP) algorithm is considered to be the state of the art method for geometric alignment of point cloud and range data which we briefly summarise in Section \ref{subsec:ICP}. ICP was first described in \cite{Besl1992} and \cite{Chen1991} and has been extensively used in scan matching and modelling \cite{Newcombe2011}\cite{Salas-Moreno2013}, due to its robustness and speed to compute. The algorithm attempts to iteratively minimise the distance between corresponding points in multiple point clouds by rigidly transforming them. 

\begin{figure}[t]
	\begin{center}
		\includegraphics[width=\linewidth]{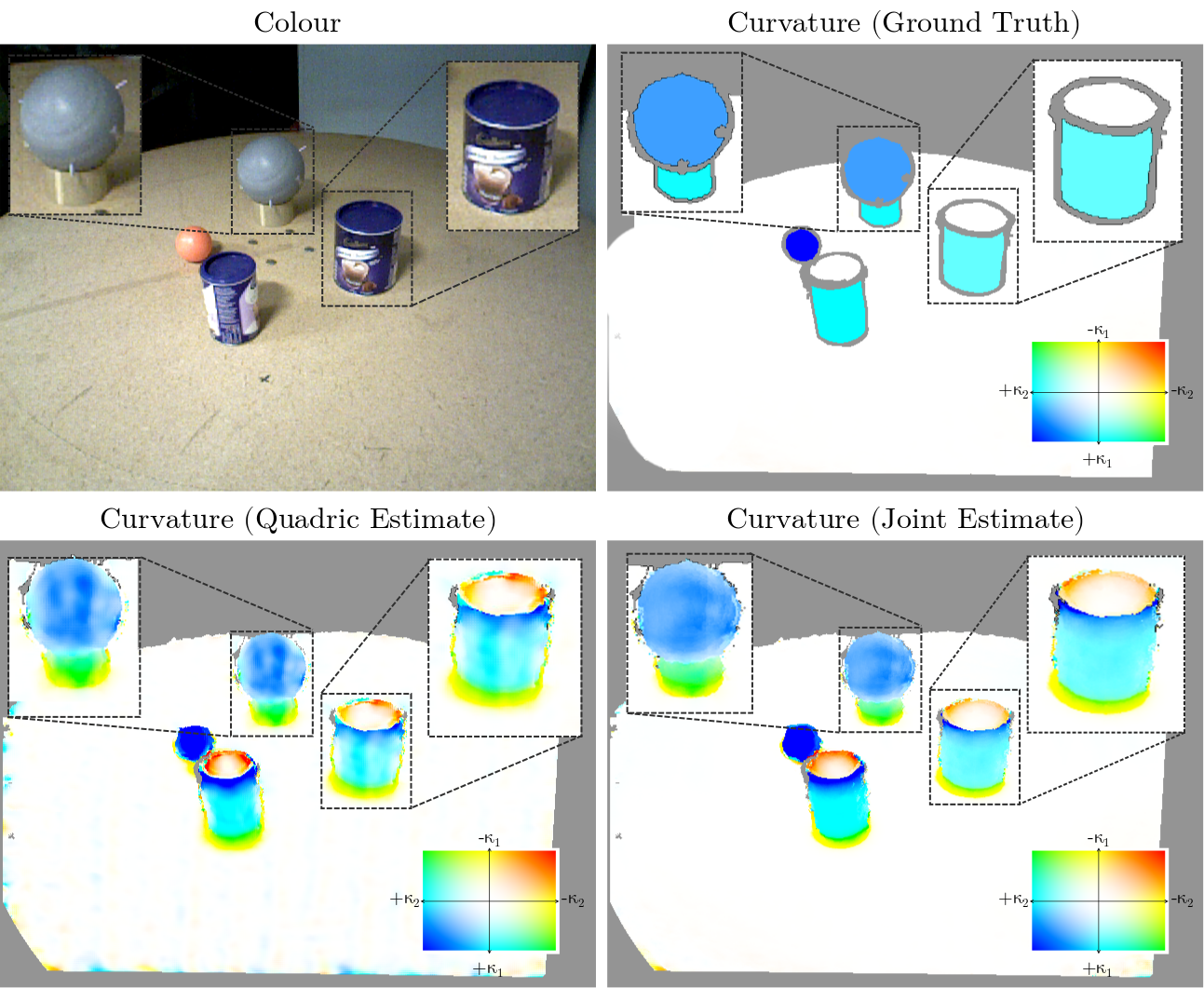}
		\caption{\emph{\textbf{Top Left:}} shows the original colour image, \emph{\textbf{Top Right:}} shows the manually labelled ground truth curvature value, \emph{\textbf{Bottom Left:}} shows the result of estimating curvature using a single frame \cite{Spek2015}, \emph{\textbf{Bottom Right:}} shows the result from this work, using the joint-full multi-frame optimisation. We enlarge the same sections of each image to provide the reader a better view of the differences. Additionally we include a key to indicate how the values of the principal curvatures ($ \kappa_1, \kappa_2 $) relate to the mapped colours, in the bottom corner of each image.}  
		\label{fig:qualitative}
	\end{center}
	\vspace{-2em}
\end{figure}

A more general solution to ICP was described in \cite{Mitra2004}, where Mitra \emph{et al.} show the increased accuracy of using quadric based scan alignment. They also demonstrate the improved convergence radius of their method over previous implementations of ICP. However, the presented methodology uses a fixed estimate of quadrics for frame-to-frame alignment which can be affected by noise in data, and no real consideration is made for online performance.

In this paper we present a novel joint optimisation approach that solves for pose alignment and curvature simultaneously using a joint quadric optimisation function (Section \ref{sec:joint_opt}). We present the results of two implementations that use our novel joint cost function, which measures how well a set of quadrics fit over multiple frames. The first novel implementation is designed to be used as an online addition to a tracking system, reducing the relative pose error in adjacent point-clouds while simultaneously improving the current dense estimates of principal curvature values (Section \ref{subsec:joint_ftf}). The second novel implementation is intended to be used offline, performing a joint pose and curvature optimisation across multiple frames after they have been captured and provided with some estimate of their global alignment (Section \ref{subsec:joint_full}). 

In Section \ref{sec:results} we examine the performance of our novel implementations compared with state of the art alternatives, and demonstrate the improved accuracy of our approaches. We also test our approach using real-world low cost depth sensor datasets to demonstrate its applicability to robotic applications. In Section \ref{subsec:pose_eval} we show our implementations produce improved pose alignment over conventional methods. While in Section \ref{subsec:pose_drift} we demonstrate that our novel approaches can be used to significantly reduce the effect of pose drift, which can be a problem for loop-detection. Finally in Section \ref{subsec:curv_eval} we demonstrate the improved accuracy of our novel implementations for curvature estimation using multiple frames.


\section{Related Work}

ICP \cite{Besl1992}\cite{Chen1991} is a modelling technique important in robotics and computer vision for navigation and mapping \cite{Salas-Moreno2013}. In \cite{Besl1992}, Besl introduced several different formulations of the minimisation problem including point-to-point, point-to-plane and plane-to-plane, which refer to the method of computing the distance between correspondences. In \cite{Rusinkiewicz}, Rusinkiewicz \emph{et al.} investigate the accuracy and speed of each formulation, finding point-to-plane to be the best choice for most applications leading to its popularity in tracking applications\cite{Newcombe2011}\cite{Newcombe2011b}\cite{Salas-Moreno2013}\cite{Whelan12rssw}. This method of point-cloud alignment is considered state of the art for aligning multiple overlapping depth scans or point clouds. 

ICP relies on largely static scenes in order to operate and can be used to generate both models and accurate ego-motion tracking. Point-to-Plane (Pt-Pl) ICP is used extensively in modern Simultaneous Localisation and Mapping (SLAM) systems  as the primary tracking algorithm. In \cite{Izadi2011} Izadi \emph{et. al.} use Pt-Pl ICP to produce a highly accurate and robust dense model, by tracking against the current model, significantly reducing pose drift. This method of tracking is fast to compute using GPUs, while providing reasonably high accuracy. However, this method of tracking against a model requires a large amount of data to be stored and a complex algorithm to move model sections in and out of memory \cite{Whelan12rssw}.

An improved pose alignment method to point-to-plane ICP was proposed in \cite{Mitra2004}, where Mitra \emph{et al.} use local surface approximations (quadrics) to provide improved pose alignment accuracy. Additionally they demonstrate an improvement in the convergence radius and increased robustness of choosing initial alignment. However, the increase in convergence radius is achieved through a simple heuristic which measures how \textit{correct} the current alignment is and randomly perturbs the current solution if it appears to be converging incorrectly. Additionally their implementation provides no real consideration to online performance of their system.

An important geometric feature used in robotics and computer vision is surface curvature. Curvature has been shown to be useful for many computer vision and robotics applications\cite{Guillaume2004}\cite{Rusinkiewicza}, primarily segmentation \cite{Besl1988} and improved modelling \cite{Rusu2010a}. In \cite{Besl1988}, Besl \emph{et al.} demonstrates that regions of high curvature often occur on object boundaries and this can be used as a reliable basis for segmentation. This idea was extended in \cite{Guillaume2004a}, where Guillaume \emph{et al.} apply this work to mesh based segmentation. However, these previous approaches either only applied to a single model \cite{Griffin2011}\cite{Rusinkiewicza}, or rely on the alignment of multiple scan-matches to highly accurate in order to build a single shape model for curvature estimation. Additionally, these previous approaches often assume relatively low-noise in the model, which is not the case when using low-cost depth sensors. In \cite{Spek2015}, Spek \emph{et al.} presents a method for real-time curvature computation that solves the problem of sensor noise using a quadric based representation. 


\section{Fundamentals}

We begin with a short revision of the methods we have used in evaluation of our system, and based our current novel implementation upon. We briefly explain ICP as used in \cite{Rusinkiewicza} and our previous work on extracting curvature from quadrics\cite{Spek2015}.

\subsection{Iterative Closest Point (ICP)}
\label{subsec:ICP}
Iterative Closest Point (ICP) is a technique for aligning two overlapping 3D scans or point clouds of a static scene and is considered state of the art in terms of accuracy and speed. The technique was first developed by Besl et at. \cite{Besl1988} and Chen et al. \cite{Chen1991}. We use a variant of ICP to provide a baseline result for the relative pose estimation, known as point-to-plane ICP \cite{Rusinkiewicz}.

This operates by defining an error function which measures the Euclidean distance from transformed points in surface 1 $(S_1)$, to a small plane patch on surface 2 $(S_2)$.

A normal is computed for every point $\bar{q} \in S_2$, using a small neighbourhood $ N(\bar{q}) $ around it. For each point $\bar{p}_i \in S_1$ a transformation estimate is applied to find the closest point $\bar{q}'_i \in S_2$ using Euclidean distance. The error is computed using the point-to-plane distance between $\bar{q}_i$ and $\bar{p}_i$ defined as
\begin{equation}
\label{eq:ptpl_error_func}
e_i = \bar{n_i}^T(R(\bar{\theta})\bar{p}_i + \bar{t} - \bar{q}_i),
\end{equation}
where 
\[T(\theta) =\left[
\begin{array}{cccc}
&  &   &  \\
\multicolumn{3}{c}{R(\theta)} & \bar{t} \\
& & &  \\ 
0 & 0 & 0 & 1
\end{array}
\right],\]
is the transformation from frame 1 to frame 2. 

The goal of ICP is to minimise the sum of the squared errors
\begin{equation}
\mathbf{C} = \sum_{i}^{N}{e_i(\bar{\theta})}^2
\end{equation}
Using a standard Gauss-Newton approach requires derivative of the error function w.r.t. the motion parameters ($\theta_i$) of the transformation $T$ giving:
\begin{equation}
J_{i} = \frac{\partial e_i}{\partial \bar{\theta}} = \left[\hat{n}^T, (\bar{q}_i\wedge\hat{n})\right]
\end{equation}
\vspace{-1em}
\begin{align}
\text{thus } \Delta\bar{\theta} = -J^T\bar{e} &= -\left[J^TJ\right]^{-1}J^T\bar{e}\\
\label{eq:least_squares_update}\text{and with re-weighting } &= -\left[J^TWJ\right]^{-1}J^TW\bar{e}
\end{align}
where W is a diagonal matrix and $w_i = f(e_i)$ and $\Delta\bar{\theta}$ is the update to the pose estimate. Equation \ref{eq:least_squares_update} gives the standard solution using re-weighted least squares. The updated pose is given by a simple matrix exponential equation, as shown in \cite{helgason1979differential}:
\begin{equation}
\label{eq:pose_update}
T_{t+1} = e^{\sum_{i=1}^{1}\theta_i G_i}T_{t}
\end{equation}

\subsection{Quadric Surface Representation}
\label{subsec:quad_surface}

As with our previous implementation in \cite{Spek2015} we compute curvature using a quadric representation, which we define as:
\begin{equation}
\label{eq:quad}
\bar{q_i}^T E^T F E \bar{q_i} = 0
\end{equation}
where
\begin{equation}
F = \left[ 
\begin{array}{cccc}
\tfrac{a}{2} & \tfrac{b}{2} & 0 & 0 \\ [4pt]
\tfrac{b}{2} & \tfrac{c}{2} & 0 & 0 \\ [4pt]
0 & 0 & 0 & -\frac{1}{2} \\ [4pt]
0 & 0 & -\frac{1}{2} & 0 \\
\end{array} 
\right]
\end{equation}
is the parabolic quadric matrix $Q$ that contains the second fundamental form $\mathbf{II}$ as the upper-left corner and
\begin{equation}
E = \left[
\begin{array}{cccc}
&  &   & 0 \\
\multicolumn{3}{c}{R(\theta_x, \theta_y)} & 0 \\
& & & t_z \\ 
0 & 0 & 0 & 1
\end{array}
\right]
\end{equation}
where $ R(\theta_x,\theta_y) $ is the Euclidean rotation matrix that rotates the local coordinate $\bar{q_i}$ to the normally aligned local frame of $F$, and $t_z$ allows the centre of the frame to sit above the central coordinate. The principal curvatures ($\kappa_1, \kappa_2$) are the Eigen values of the second fundamental form and can be extracted directly from $F$ using:
\begin{equation}
\begin{aligned}[c]
t_1 &= \frac{a+c}{2}\\
t_2 &= \sqrt{t_1^2 - ac + b^2}\\
\end{aligned}
\hspace{1em}
\begin{aligned}[c]
\kappa_1 &= t_1+t_2\\
\kappa_2 &= t_1-t_2\\
\end{aligned}
\end{equation}

We iteratively fit the quadric to the surface by re-expressing Equation \ref{eq:quad} as an error function:
\begin{equation}
\label{eq:quad_error_func}
\epsilon_i(\bar{\gamma}_p) = \bar{q}_i^T Q(\bar{\gamma_p}) \bar{q}_i
\end{equation}
where $Q(\bar{\gamma_p})=E^T F E$ is the quadric defined at the central point $\bar{p}$ of the quadric neighbourhood and $\bar{\gamma_p}=\{\theta_x, \theta_y, t_x, a, b, c\}$ are the six parameters of the quadric. We use $\epsilon_i$ to define a cost function for the points in the neighbourhood of $\bar{q}_i \in N(\bar{p})$ around the central point
\begin{equation}
\label{eq:quad_cost_func}
\mathbf{C}(\bar{\Gamma}) = \sum_{\bar{p} \in S}^{} \sum_{\bar{q}_i \in N(\bar{p})}^{} \epsilon_i(\bar{\gamma}_p)^2
\end{equation}
which we minimise using a standard Gauss-Newton approach shown in Equation \ref{eq:least_squares_update} to solve for the quadric parameters ($\bar{\Gamma}$) that minimise the cost function. Where $\bar{\Gamma} = \{\bar{\gamma}_p\}$, the set of quadric parameters for each point in the surface $S$ and $N(p)$ is the set of points in the neighbourhood of $\bar{p}$ the central coordinate of the quadric. 

\begin{figure*}[ht!]
	\centering
	\includegraphics[width=\textwidth]{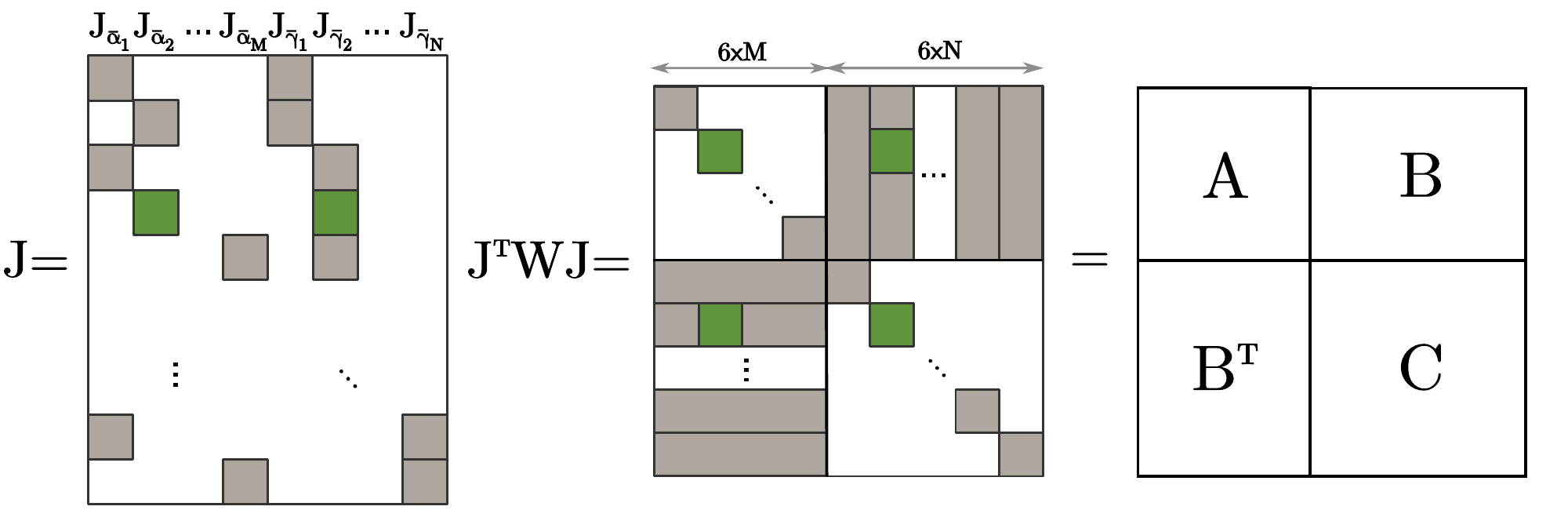}
	\vspace{1em}
	\caption{\textbf{Left:} The joint Jacobian matrix ($ J $), demonstrates the relationship between the pose ($J_{\bar{\alpha}}$) and quadric ($J_{\bar{\gamma}}$) Jacobians. Each block indicates a corresponding patch in frame $j$ for quadric $p$. \textbf{Center/Right:} The Hessian matrix ($ J^TWJ $) for the joint optimisation, results in two block diagonal matrices ($A$ and $C$) and a potentially dense joint information matrix ($B$). The green blocks indicate which parts a single correspondence would effect in the Jacobian and Hessian Matrix.}
	\label{fig:joint_jacobian}
\end{figure*}
\section{Joint Optimisation}
\label{sec:joint_opt}

In this section we detail the novel extension of our previous quadric error function (see above) to allow the joint optimisation for pose and principal curvature values.

\subsection{Frame-to-Frame}

We now extend the quadric error function from \cite{Spek2015} to include an additional frame:
\begin{equation}
\label{eq:joint_ftf_error}
\begin{split}
\epsilon_p(\bar{\alpha},\bar{\gamma_p}) &= \sum_{\bar{p}_k \in N(\bar{p}, S_1)}^{} \left(\bar{p}_k^TQ(\bar{\gamma_p})\bar{p}_k\right)^2\\ 
&+ \sum_{\bar{q}_k \in N(T(\bar{\alpha}_j)^{-1}\bar{p}, S_2)}^{} \left(\bar{q}_k^TT(\bar{\alpha}_j)^TQ(\bar{\gamma_p})T(\bar{\alpha}_j)\bar{q}_k\right)^2 
\end{split}
\end{equation}
This adds the additional motion parameters $\bar{\alpha}$ of the relative pose alignment of the second surface to the original. The first part of the error function is as before while the second half adds the information from the additional frame. For the second frame we take the neighbourhood ($\bar{q}_k \in N(T^{-1}p, S_2)$) around the quadric central point $\bar{p}$ as it transforms into the second surface.

\subsection{Multi-Frame}

We can now trivially generalise the error function in Equation \ref{eq:joint_ftf_error} to work for the multiple frames. The error of any particular neighbourhood around a transformed central point in surface $S_j$ is given by:
\begin{equation}
\label{eq:joint_error}
\epsilon_p(\bar{\alpha_j},\bar{\gamma}) = \sum_{\bar{q}_k \in N(\bar{q}, S_j)}^{} \left(\bar{q}_k^TT(\bar{\alpha_j})^TQ(\bar{\gamma_p})T(\bar{\alpha_j})\bar{q}_k\right)^2
\end{equation}
where the neighbourhood is now defined around the transformed central point $ \bar{p} \in S_1 $ into the surface $S_j$ by $\bar{q}=T(\bar{\alpha_j})^{-1}\bar{p}$, where $ T_1 = \mathbf{I} $ the identity. This new error function also contains a set of motion parameters $\bar{\alpha}$ for each pose $T(\bar{\alpha})$. We can now define a joint cost function over all points $\bar{p} \in S_1$
\begin{equation}
\label{eq:joint_cost}
\mathbf{C}(\bar{A},\bar{\Gamma}) = \sum_{\bar{p} \in S_1}^{} \sum_{\bar{\alpha}_j \in \bar{A}}^{} \epsilon_p(\bar{\alpha_j},\bar{\gamma_p})
\end{equation}
where  
$
\bar{A} = \left\{\bar{\alpha}_1, \bar{\alpha}_2 \dots \bar{\alpha}_M \right\}$ and $\bar{\Gamma} = \left\{\bar{\gamma}_1, \bar{\gamma}_2 \dots \bar{\gamma}_N \right\}
$. This means a separate set of motion parameters for each surface $S_j$ and separate set of quadric parameters for every point $\bar{p} \in S_1$.

To compute parameters $\bar{A}$ and $\bar{\Gamma}$ that minimise the cost function we use a Gauss-Newton approach to iteratively reduce the error. We simplify the differentiation process for ourselves by using the chain rule. Differentiating Equation \ref{eq:joint_error} w.r.t. the motion parameters of that surface $S_j$'s pose $ \bar{\alpha_j} $ at any point in the $\bar{q'} \in N(\bar{q})$ gives the following derivative:
\begin{equation}
\label{eq:motion_derivative}
J_{\alpha_{ji}} = \bar{q}'^TT_j^TG_i^TQ(\bar{\gamma}_p)T_j\bar{q}' + \bar{q}'^TT_j^TQ(\bar{\gamma}_p)G_iT_j\bar{q}' 
\end{equation} 
where transformation $T_j$ can be expressed as a matrix exponential as shown in Equation \ref{eq:pose_update}, which allows simple application of the chain rule for differentiation. The matrix $G_i$, known as the generator, is a linear approximation of the derivative of the transformation w.r.t. the motion parameter $\alpha_{ji}$, where $\bar{\alpha_j}=\{ \theta_x,\theta_y,\theta_z,t_x,t_y,t_z\}$. This gives the full Jacobian of the motion parameters: 
\begin{equation}
\label{eq:motion_jacobian}
J_{\bar{\alpha}_{j}} = \{J_{\alpha_{j1}}, J_{\alpha_{j2}}, ... J_{\alpha_{j6}}\} 
\end{equation}

\newcommand*\rfrac[2]{{}^{#1}\!/_{#2}}

The derivative of the joint error function w.r.t. each quadric parameter is given by
\begin{equation}
\label{eq:quadric_derivative}
J_{\bar{\gamma}_{pi}} = \bar{q}^TT_j^T\frac{\partial Q_p}{\partial\gamma_{pi}} T_j\bar{q},
\end{equation}
where $\rfrac{\partial\mathbf{Q}}{\partial\gamma_{pi}}$ is the same as per the original formulation in \cite{Spek2015} (see Section \ref{subsec:quad_surface}). This gives the full Jacobian of the quadrics parameters for every point $ \bar{p} $
\begin{equation}
\label{eq:quadric_jacobian}
J_{\bar{\gamma}_{p}} = \{J_{\gamma_{p1}}, J_{\gamma_{p2}}, ... J_{\gamma_{p6}}\} .
\end{equation}

For each point in each transformed neighbourhood around point $\bar{p}$ in each surface $S_j$ we compute both Jacobians, $J_{\bar{\alpha}_j}$ and $J_{\bar{\gamma}_p}$. We can combine the measurements for all points into a single Jacobian matrix $\mathbf{J}=\{J_{\bar{\alpha}}, J_{\bar{\gamma}}\}$ as shown in Figure \ref{fig:joint_jacobian}, which demonstrates the relationship between the Jacobians for each point and the weighted Hessian matrix ($\mathbf{J}^T\mathbf{W}\mathbf{J}$). We use the standard method for computing the update to the motion and quadric parameters using Gauss-Newton. This gives
\begin{equation}
\left[
\begin{array}{c}
\Delta\bar{\alpha}\\
\Delta\bar{\gamma}
\end{array}
\right]=
\left(J^{T}WJ\right)^{-1}J^TW\bar{\epsilon},
\end{equation}
where $ \Delta\bar{\alpha} $ is the update to each pose estimate, $ \Delta\bar{\gamma} $ is an vector containing and an update to each quadric in $S_1$ and $W$ is a diagonal matrix of weights of the form $ w_i = f(\epsilon_i, d_i) $. The weighting function $ f $ we use is a modified 'fair' weight function \cite{m_est} that uses the quadric error ($ \epsilon_i $) and the distance from the central point ($ d_i $), such that
\begin{equation}
f(\epsilon_i, d_i) = \frac{n^2}{n^2 + \epsilon_i^2 + d_i^2},
\end{equation}
where $n$ is a constant that is close to the upper limit of a typical error value, and can be found experimentally.
\subsection{Schur-Complement Trick}
\label{subsec:schur_comp}
In order to solve for the update we will need invert the Hessian matrix ($ J^TWJ $). However, because we compute an update densely for all quadrics the Hessian is too large to practically invert in a reasonable amount of space and time, for example using double precision at full resolution (640x480) the Hessian matrix would consume ~27 tera-bytes of memory. In order to avoid this issue we exploit the sparsity and symmetry of the matrix and use the block diagonal Schur complement trick \cite{Ouellette1981}. This is applicable to problems of the form\begin{equation}
\label{eq:schur_complement_form}
\left[
\begin{array}{c}
\bar{x} \\
\bar{y} \\
\end{array}
\right]
= 
\left[
\begin{array}{cc}
A & B \\
B^T & C \\
\end{array}
\right]^{-1}
\left[
\begin{array}{c}
\bar{a} \\
\bar{b} \\
\end{array}
\right].
\end{equation}
This gives the block-diagonal symmetric matrix $C$ which is cheaply piece-wise invertible and $A$ which is also symmetric and invertible which is the case as shown in Figure \ref{fig:joint_jacobian}. We can rearrange this formulation to solve for the updates to the poses ($\bar{x}$) and quadrics ($\bar{y}$):
\begin{equation}
\label{eq:schur_complement_trick}
\left[
\begin{array}{cc}
A & B \\
B^T & C \\
\end{array}
\right]^{-1}
=
\left[
\begin{array}{cc}
G & -G^{-1}H\\
-H^TG^{-1} & C^{-1} + H^TGH\\
\end{array}
\right]
\end{equation}
\begin{equation*}
G = A - BC^{-1}B^T \hspace{2em} H = BC^{-1}
\end{equation*}
Giving the solution for $\bar{x}$ and $\bar{y}$ as:
\begin{equation}
\label{eq:schur_comp_update}\textsl{}
\begin{aligned}
\bar{x} &= G^{-1}(\bar{a}-H\bar{b}) \\
\bar{y} &= C^{-1}\bar{b} - H^T\bar{x} \\ 
\end{aligned}
\end{equation}

One issue with this is the solution to $\bar{x}$ must be found first in order to solve for $\bar{y}$, as indicated in Equation \ref{eq:schur_comp_update}. In this case $\bar{x}$ is an update vector for each pose estimate via a matrix exponentiation and $\bar{y}$ is a vector of updates, one per quadric. This means a two pass computation must be performed, one to solve for the pose updates and then adding this information to the quadric updates.


\section{Novel Implementations}
\label{sec:method}

We now discuss two novel implementations based on the joint error function described in section \ref{sec:joint_opt}.

\subsection{Joint Frame-to-Frame}
\label{subsec:joint_ftf}
Using the joint frame-to-frame cost function for the special case of two frames described in section \ref{subsec:joint_ftf} we implemented a novel joint frame-to-frame (referred to as \emph{J-ftf} in Section \ref{sec:results}) optimiser that uses two frames at a time. This novel implementation was made using CUDA and is intended to be used as on online addition to an existing tracker. The reason two-frames can be considered a special case computationally is that the B matrix from Equation \ref{eq:schur_complement_form} has only a single joint information matrix per point, which means it can be computed on the fly allowing for greater optimisation on the GPU, for further details please refer to the available implementation\footnote{Source code available from the following repository: \\https://github.com/aspek1/JointCurvatureOptimisation.git}. The input to this method is an estimate of the pose and a dense set of quadrics provided by our previous implementation \cite{Spek2015}. A key consideration of this implementation was time, so considerable effort was made to reduce execution time despite the size and complexity of the computation. Our implementation runs in approximately 125ms per iteration on an NVidia Titan X (Maxwell), allowing it to run in conjunction with an existing tracker to reduce pose drift (see Section \ref{subsec:pose_drift}) and improve principal curvature estimation (see Section \ref{subsec:curv_eval}).

\subsection{Joint Full}
\label{subsec:joint_full}
We also present a novel joint optimisation over multiple frames, which we call Joint Full (referred to as \emph{J-full} in Section \ref{sec:results}). This is a post-process step implemented on CPU to run on a dataset and improve the relative pose estimates and curvature for each surface. The implementation is very similar to that for the frame-to-frame version (see previous), the main difference is that we compute the absolute pose estimates for all frames simultaneously. For each surface we compute an estimate of the transformed point $\bar{p}$ into the surface, giving us a set of correspondences. Now we compute pose and quadric Jacobians for each correspondence, and update an the appropriate Jacobian Matrix as indicated in Figure \ref{fig:joint_jacobian}. Finally we compute the update for each pose and quadric using Schur's Complement as described in Section \ref{subsec:schur_comp}.


\begin{figure}[h!]
	\begin{center}
		\includegraphics[width=\linewidth]{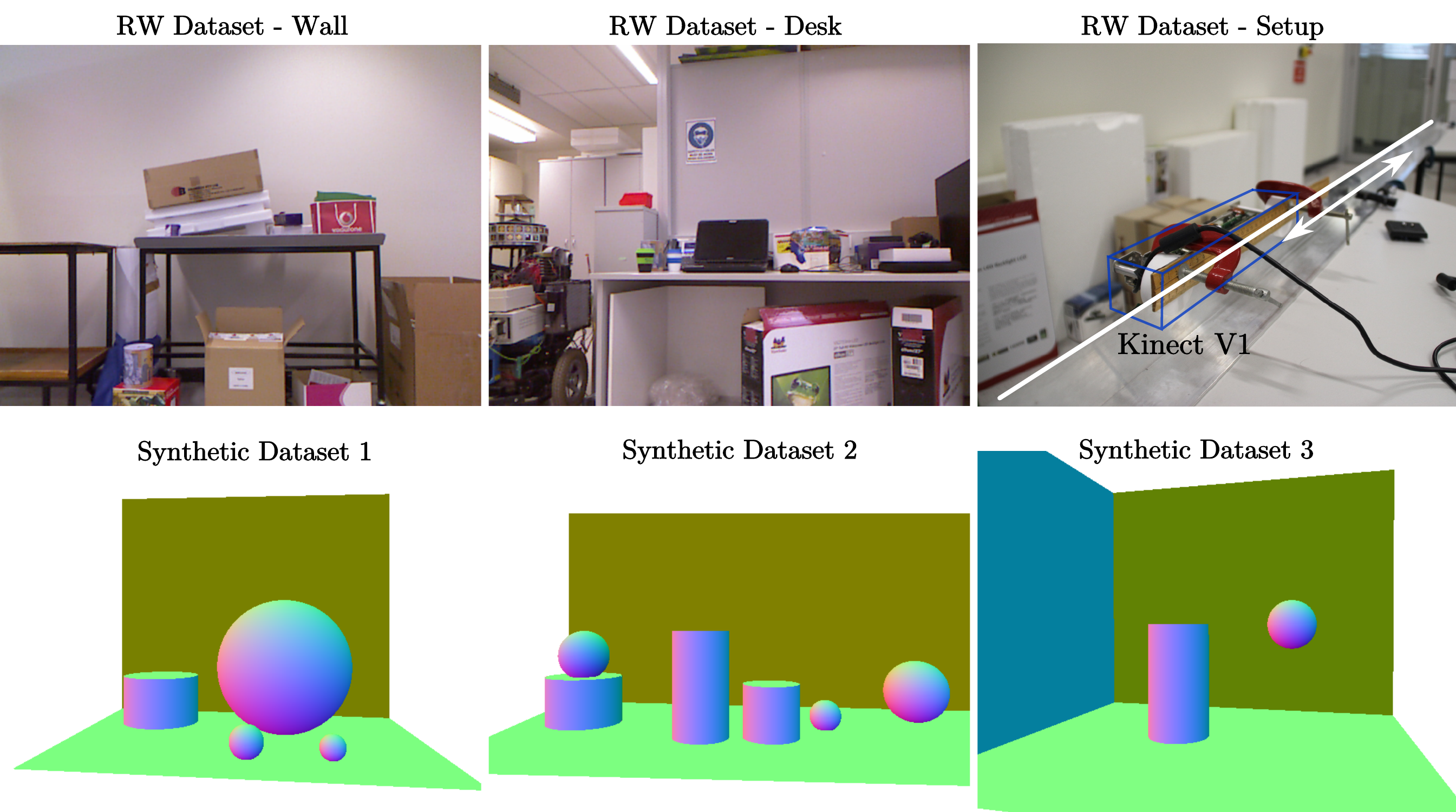}
		\caption{The synthetic and real-world datasets used to assess pose accuracy of several approaches. \emph{\textbf{Top-Right:}} Demonstrates the experimental setup used to capture the real-world datasets \emph{Wall} and \emph{Desk}. A Microsoft Kinect V1 is attached to a rail and moved in rigidly along a straight-line trajectory capturing frames every 10mm.} 
		\label{fig:pose_dataset}
	\end{center}
	\vspace{-1.5em}
\end{figure}

\begin{figure}[h!]
	\begin{center}
		\includegraphics[width=\linewidth]{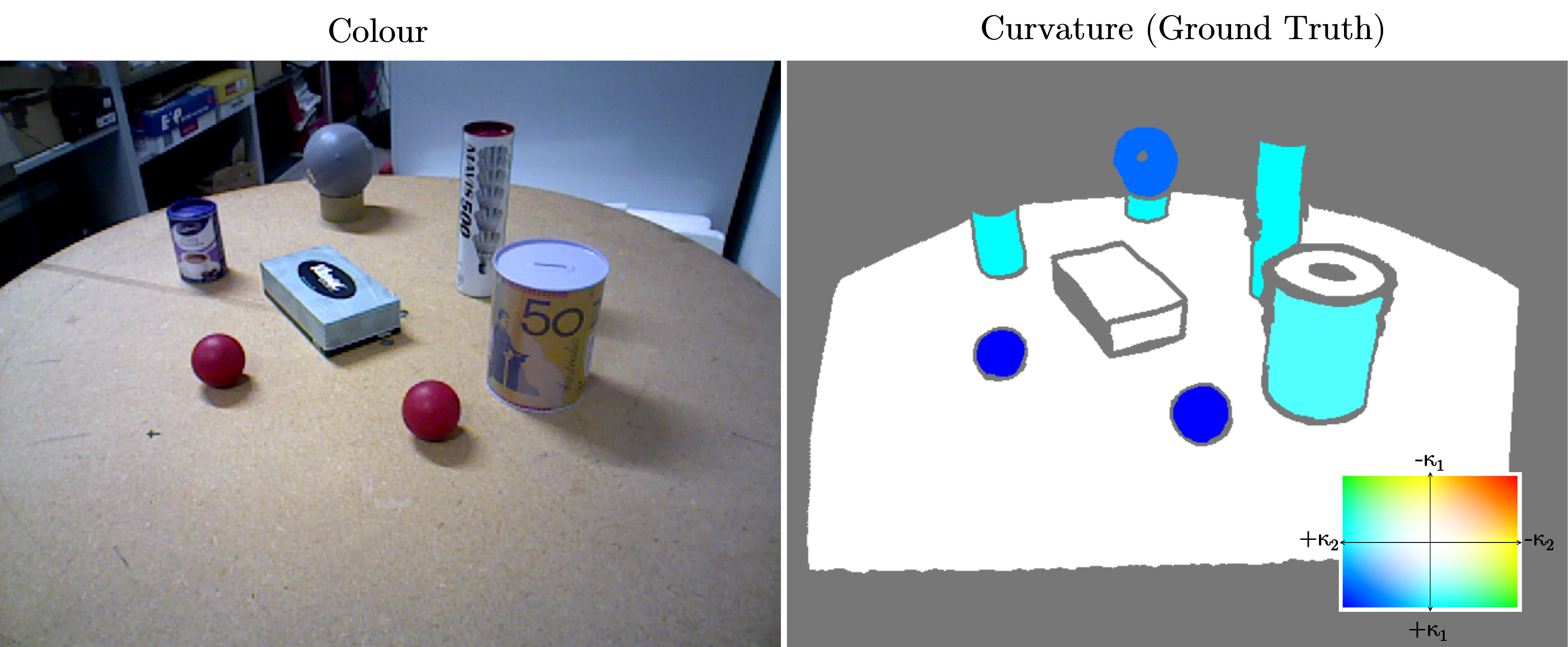}
		\caption{An example from one of the real-world curvature datasets (\emph{Real-World 5}). \emph{\textbf{Right:}} the colour image, \emph{\textbf{Left:}} manually labelled ground-truth curvature values. } 
		\label{fig:curv_dataset}
	\end{center}
	\vspace{-1.5em}
\end{figure}
\section{Results}
\label{sec:results}

\subsection{Testing Methodology}
\label{sec:testing_methodolgy}

In Section \ref{subsec:pose_eval} we compare the performance of our system to several state of the art algorithms for pose alignment on synthetic and real-world ground truth datasets. The synthetic datasets are generated using ray-tracing with Gaussian noise successively added as a percentage of the distance to the camera to examine the robustness of each method to noise. The two real-world datasets are on-rails, translation only datasets, where a frame is captured every 10mm of horizontal motion by a Microsoft Kinect as shown in Figure \ref{fig:pose_dataset}. 

The tested pose alignment methods include; our joint frame-to-frame (\emph{J-ftf}) approach (see Section \ref{subsec:joint_ftf}) and our full joint optimisation (\emph{J-full}) (see Section \ref{subsec:joint_full}), a dense variant of frame-to-frame point-to-plane (Pt-Pl) ICP (\emph{ICP-ftf}) \cite{Rusinkiewicz} and a dense Pt-Pl ICP bundle adjustment (\emph{ICP-bundle}) \cite{Triggs2000}, and a global pose alignment to a single set of quadrics (\emph{Q-full}) similar to J-full. Unlike \emph{J-full} approach \emph{Q-full} only optimises for the absolute pose updates, maintaining a constant estimate of the original quadrics from \cite{Spek2015}. The Pt-Pl ICP bundle adjustment optimises over semi-dense a pose graph, where the weighted error across all poses is reduced jointly.

In Section \ref{subsec:pose_drift} we examine the same pose alignment methodologies as in Section \ref{subsec:pose_eval} but in terms of pose drift on real-world datasets. 

In Section \ref{subsec:curv_eval} we examine the curvature estimation accuracy for several approaches, using synthetic and real-world datasets. The synthetic datasets used are the same as in Figure \ref{fig:pose_dataset}. The real-world datasets are manually labelled with ground-truth curvature values as shown in Figure \ref{fig:curv_dataset}.  We compare the results of our joint approaches to a common curvature estimation quadric based least-squares approach (\emph{Quad LS}) from \cite{Douros2002} and our previous method iterative approach (\emph{Quad IT}) from \cite{Spek2015}. 

\begin{table}[h]
	\centering
	
	\resizebox{\linewidth}{!}{%
		\begin{tabular}{cccccc}
			\multicolumn{2}{c}{Translational Error (m)}\\ \midrule
			\textit{Noise Added}     & ICP-ftf & ICP-bundle & Ours (Q-full) & Ours (J-ftf) & Ours (J-full) \\ \midrule
			$\sigma$  & 4.19e-3 & 6.93e-4    & \textbf{2.79e-4}      & 2.87e-4   & 2.88e-4    \\
			$1\sigma$ & 9.22e-3 & 1.83e-3    & 8.15e-4      & 1.02e-3   & \textbf{6.90e-4}    \\
			$2\sigma$ & 1.52e-2 & 3.00e-3    & 1.68e-3      & 1.92e-3   & \textbf{1.53e-3}    \\
			$3\sigma$ & 2.01e-2 & 4.45e-3    & 2.53e-3      & 2.91e-3   & \textbf{2.37e-3}   \\ \midrule
			\multicolumn{2}{c}{Rotational Error (rad)} \\ \midrule
			$\sigma$  & 3.36e-3 & 5.77e-4    & 2.38e-4      & \textbf{2.31e-4}   & 2.79e-4    \\
			$1\sigma$ & 7.35e-3 & 1.69e-3    & \textbf{3.69e-4}      & 5.05e-4   & 4.98e-4    \\
			$2\sigma$ & 1.31e-2 & 2.69e-3    & 6.84e-4      & 8.70e-4   & \textbf{5.54e-4}    \\
			$3\sigma$ & 1.72e-2 & 3.40e-3    & 1.21e-3      & 1.22e-3   & \textbf{8.75e-4}   \\ \midrule
		\end{tabular}%
	}
	\caption{Synthetic Pose Error - Dataset1}
	\label{tb:synth_pose_err_ds1}
\end{table}

\begin{table}[h]
	\centering
	\resizebox{\linewidth}{!}{%
		\begin{tabular}{cccccc}
			\multicolumn{2}{c}{Translational Error (m)}\\ \midrule
			\textit{Noise Added}     & ICP-ftf & ICP-bundle & Ours (Q-full) & Ours (J-ftf) & Ours (J-full) \\ \midrule
			$\sigma$  & 3.56e-3 & 5.42e-5    & \textbf{3.55e-5 }     & 3.69e-5   & 3.58e-5    \\
			$1\sigma$ & 4.83e-3 & 6.22e-4    & 4.74e-4      & 4.70e-4   & \textbf{4.56e-4}    \\
			$2\sigma$ & 8.29e-3 & 1.41e-3    & 1.07e-3      & 9.97e-4   & \textbf{9.86e-4}    \\
			$3\sigma$ & 1.17e-2 & 2.60e-3    & 2.07e-3      & 2.07e-3   & \textbf{1.97e-3}   \\ \midrule
			\multicolumn{2}{c}{Rotational Error (rad)} \\ \midrule
			$\sigma$  & 2.00e-4 & 2.71e-5    & \textbf{1.14e-5}      & 1.19e-5   & 8.49e-5    \\
			$1\sigma$ & 2.46e-3 & 1.01e-4    & 6.33e-5      & \textbf{5.92e-5}   & 9.75e-5    \\
			$2\sigma$ & 5.70e-3 & 5.34e-4    & 3.54e-4      & \textbf{2.01e-4}   & 2.10e-4    \\
			$3\sigma$ & 7.79e-3 & 2.32e-3    & 1.99e-4      & 2.26e-4   & \textbf{1.89e-4}    \\ \midrule
		\end{tabular}%
	}
	\caption{Synthetic Pose Error - Dataset2}
	\label{tb:synth_pose_err_ds2}
\end{table}

\begin{table}[h!]
	\centering
	\resizebox{\linewidth}{!}{%
		\begin{tabular}{cccccc}
			\multicolumn{2}{c}{Translational Error (m)}\\ \midrule
			\textit{Noise Added}     & ICP-ftf & ICP-bundle & Ours (Q-full) & Ours (J-ftf) & Ours (J-full) \\ \midrule
			$\sigma$  & 1.94e-3 & 1.27e-4    & 1.91e-4      & 1.23e-4   & \textbf{1.22e-4 }   \\
			$1\sigma$ & 4.61e-3 & 9.78e-4    & 9.49e-4      & 9.45e-4   & \textbf{9.39e-4}    \\
			$2\sigma$ & 8.50e-3 & 1.95e-3    & 1.96e-3      & \textbf{1.82e-3}   & 1.83e-3    \\
			$3\sigma$ & 1.58e-2 & 3.04e-3    & \textbf{2.69e-3}      & 2.71e-3   & 2.72e-3   \\ \midrule
			\multicolumn{2}{c}{Rotational Error (rad)} \\ \midrule
			$\sigma$  & 1.93e-3 & \textbf{7.09e-5}    & 1.74e-4      & 7.64e-5   & 7.40e-5    \\
			$1\sigma$ & 4.74e-3 & 2.28e-4    & 2.73e-4      & \textbf{2.55e-4}   & \textbf{2.55e-4}    \\
			$2\sigma$ & 8.25e-3 & 3.82e-4    & 4.99e-4      & 4.10e-4   & \textbf{3.69e-4}    \\
			$3\sigma$ & 1.30e-2 & 7.06e-4    & 5.52e-4      & 5.99e-4   & \textbf{5.30e-4}    \\ \midrule
		\end{tabular}%
	}
	\caption{Synthetic Pose Error - Dataset3}
	\label{tb:synth_pose_err_ds3}
\end{table}
\subsection{Pose Estimation Accuracy}
\label{subsec:pose_eval}

\subsubsection{Synthetic Dataset Evaluation}

We examine pose alignment accuracy on our synthetic datasets by measuring the RMS translation and rotation error for each of the methods described in Section \ref{sec:testing_methodolgy}, presenting the results in Tables \ref{tb:synth_pose_err_ds1},\ref{tb:synth_pose_err_ds2} and \ref{tb:synth_pose_err_ds3}. We highlight the lowest error in each category using \textbf{bold} formatting. We observe a consistent order of magnitude improvement over \emph{ICP-ftf} by our \emph{J-ftf} approach. Which is significant as our process can be used to massively improve the quality of frame-to-frame alignments during tracking. We also observe a modest improvement for our \emph{J-full} system over a full bundle adjustment (\emph{ICP-bundle}) (approximately 30\%-50\%), despite using an approach that only aligns to a single frame. This is a significant result as \emph{ICP-bundle}, uses the information of all frame-to-frame alignments to compute a semi-dense pose graph. More surprising is that our \emph{J-ftf} approach is able to consistently out-perform the \emph{ICP-bundle} approach for synthetic data.

\begin{figure}[h!]
	\begin{center}
		\includegraphics[width=\linewidth]{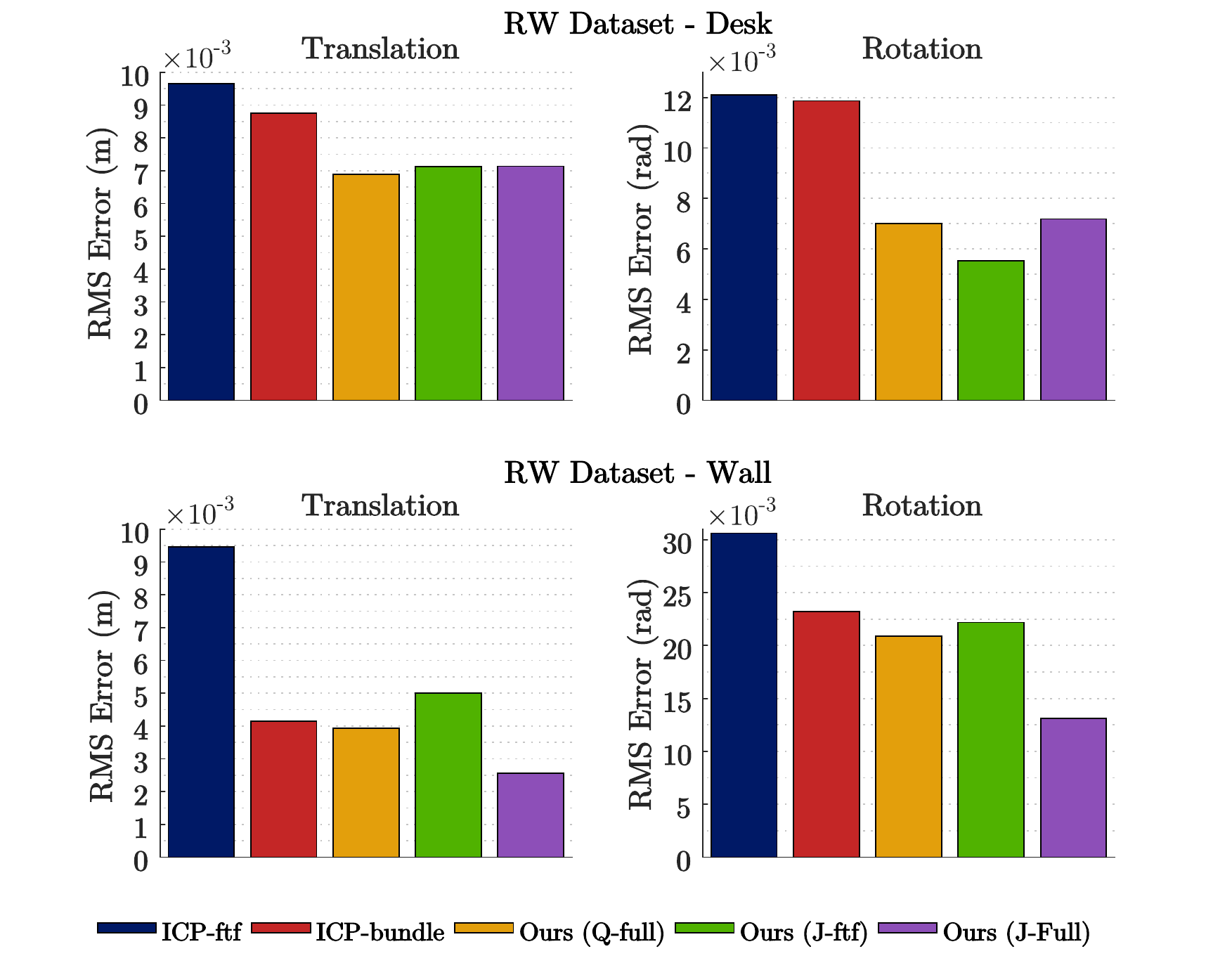}
		\caption{Real-world pose error on two ground truth datasets. The datasets were both captured using an on-rails setup with translation only along a straight trajectory. These results demonstrate the improved accuracy of our methods over the standard approaches particularly in the Dataset \emph{Desk (top)}, which shows a noticeable reduction in translation and rotational error for the joint approaches.}
		\label{fig:realworld_pose_error}
	\end{center}
	\vspace{-2em}
\end{figure}

\subsubsection{Real-world Dataset Evaluation}
We now examine the performance of our methods on real-world datasets collected using a low-cost depth sensor (the Microsoft Kinect). As shown in Figure \ref{fig:realworld_pose_error} we observe a similar reduction in pose estimation error by using our approaches, compared to the dense ICP based approaches. In dataset \emph{Wall} we see a significant improvement by our \emph{J-full} approach over other approaches, despite the large number of planar surfaces visible in the scene, which should help Pt-Pl ICP. In dataset \emph{Desk} the improvement is less significant for our methods, although all quadric based approaches significantly reduce rotation error. This evaluation is made difficult by the structured noise in the sensor, which we attempted to reduce through a device calibration.

\begin{figure}[h!]
	\begin{center}
		\includegraphics[width=\linewidth]{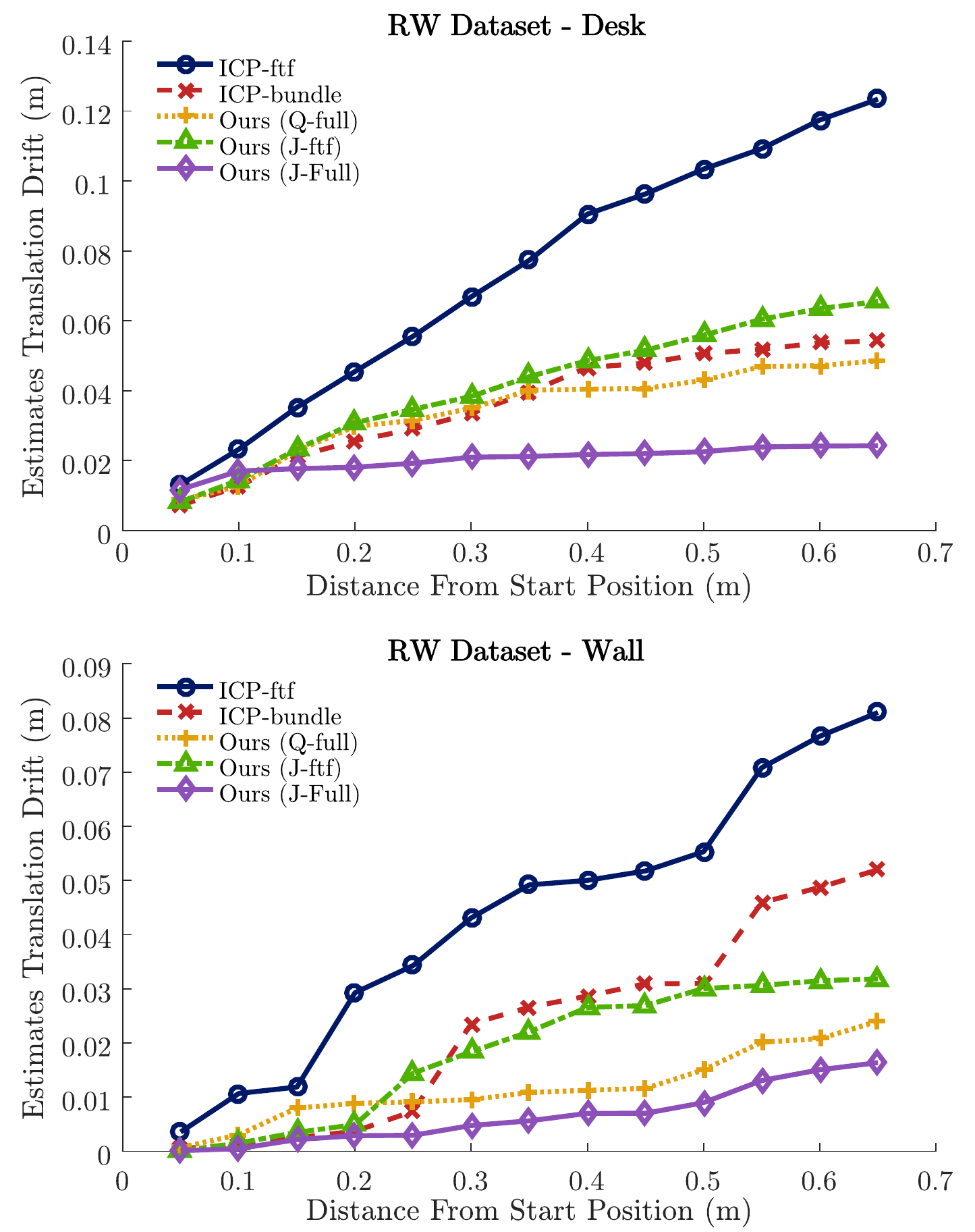}
		\caption{Demonstrates the reduction in drift using our methods, including our frame-to-frame method which even manages to match the drift of the global ICP bundle adjustment.}
		\label{fig:rw_drift}
	\end{center}
	\vspace{-1em}
\end{figure}
\subsection{Pose Estimation Drift}
\label{subsec:pose_drift}
In order to demonstrate the applicability of our systems to real-world tracking applications we examine the estimated pose drift as we move along a known trajectory. As shown in Figure \ref{fig:rw_drift} we observe a significant reduction in frame-to-frame drift using quadric based approaches. Most surprising is the reduction in drift using our \emph{J-ftf} implementation, achieving comparable pose drift error to \emph{ICP-bundle} and greatly improving upon the drift of \emph{ICP-ftf}. This shows our implementation can be used in a tracking system to greatly reduce the drift of adjacent keyframes, which can accumulate in a system and make the detection of loops very difficult for a non-feature based approach. Additionally our \emph{J-full} approach shows a significant reduction, and increases significantly slower than all other approaches. The main reason for this is that methods like \emph{ICP-bundle} benefit greatly from large loop-closures, which are not present unless the camera revisits a location.

\begin{figure}[h!]
	\begin{center}
		\includegraphics[width=.97\linewidth]{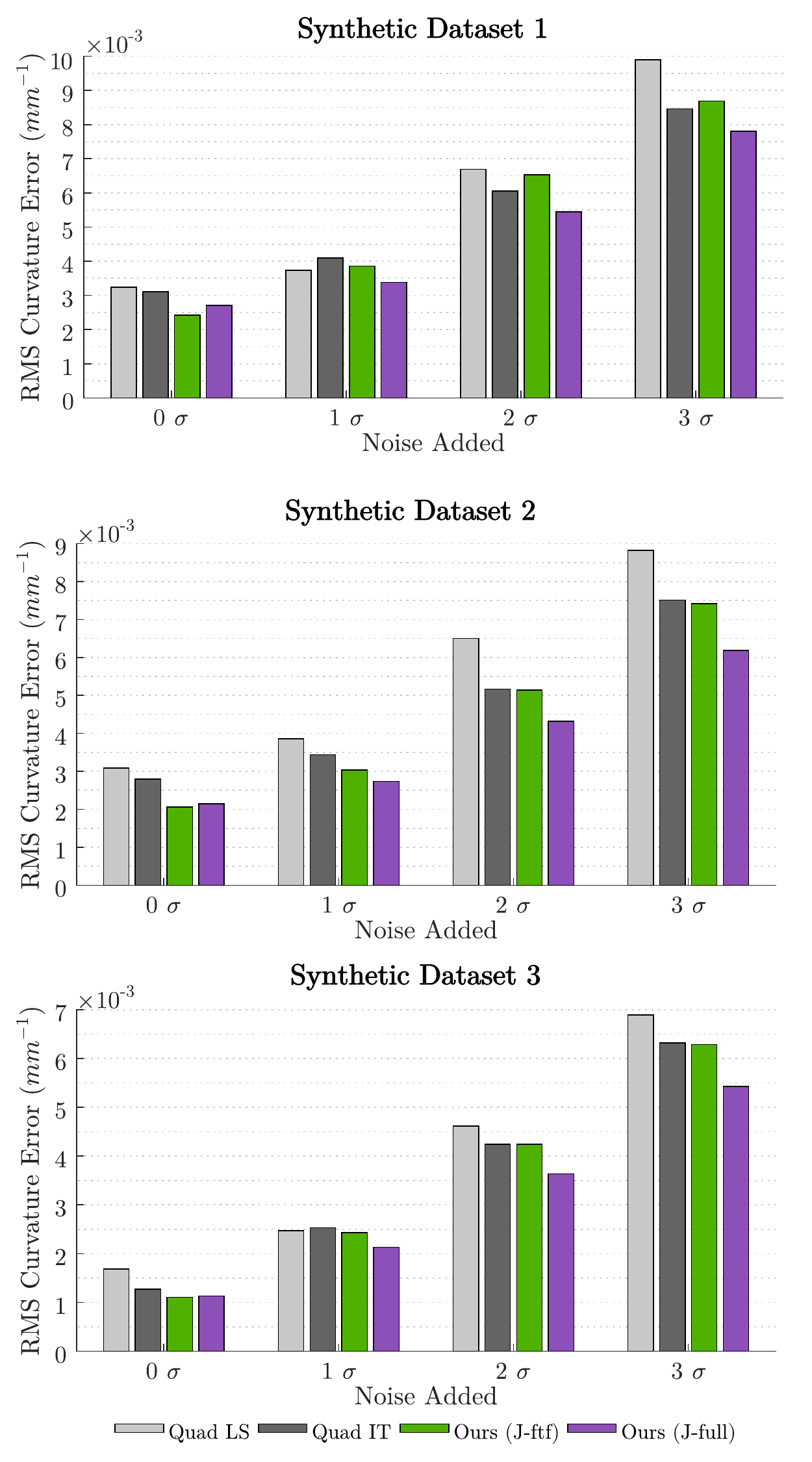}
		\caption{Results of testing several curvature estimation methods on 3 synthetic datasets with successively increasing levels of noise added to each. Each $ \sigma $ represents a 1\% Gaussian noise addition. This demonstrates our systems ability to cope with aggressive noise, and shows the improvement upon previous curvature estimation methods by using multiple frames.}
		\label{fig:curv_err_synth}
	\end{center}
	\vspace{-1em}
\end{figure}

\begin{figure}[h!]
	\begin{center}
		\includegraphics[width=\linewidth]{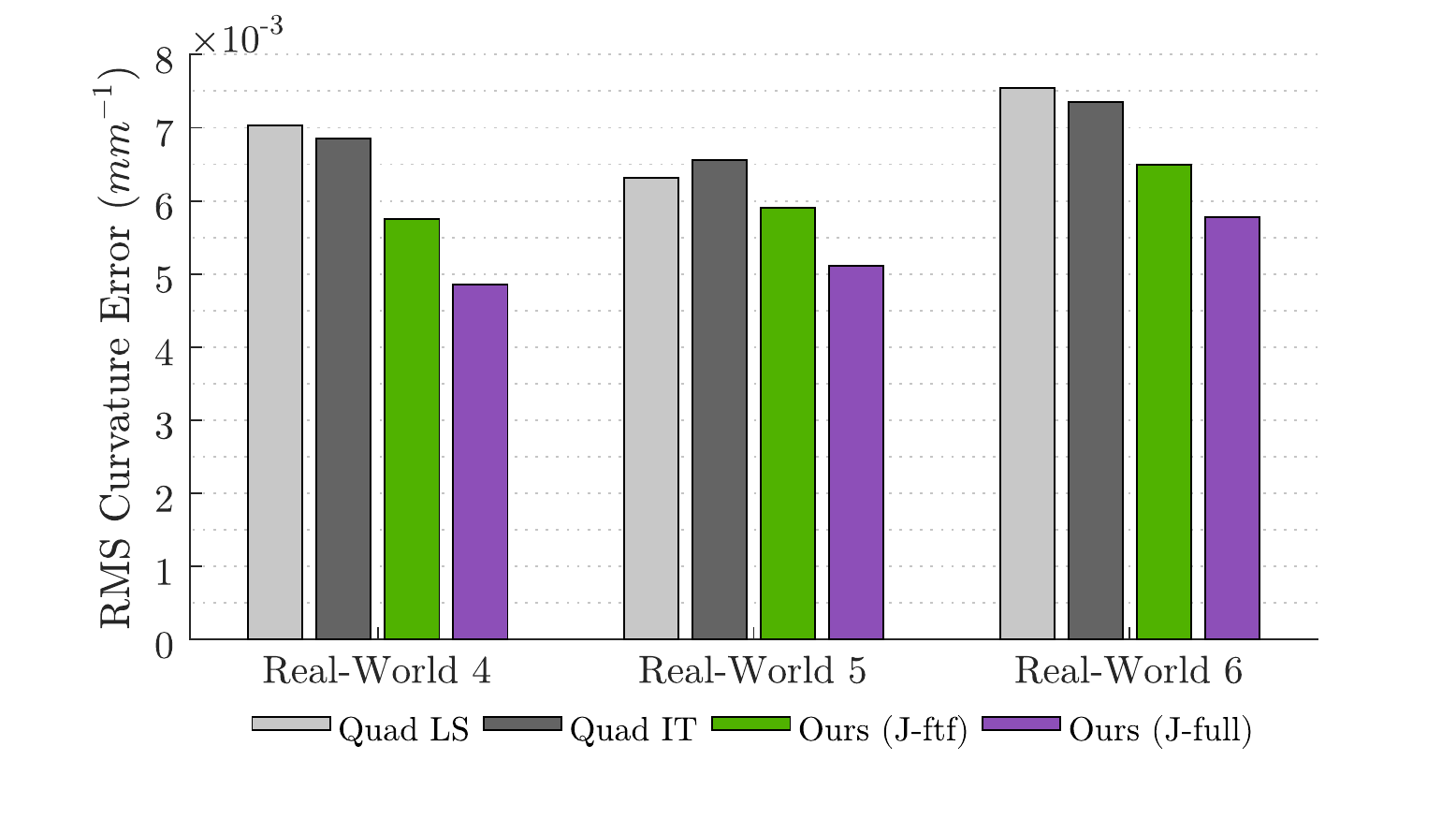}
		\caption{RMS curvature error for 3 real-world manually labelled datasets. Note the reduction in error upon previous state-of-the-art techniques across all datasets using our joint approaches. Even with just two frames (\emph{J-ftf}) our system shows a significant reduction in error.}
		\label{fig:curv_err_real}
	\end{center}
\end{figure}

\subsection{Curvature Estimation Improvement}
\label{subsec:curv_eval}

\subsubsection{Synthetic Dataset Evaluation}
We examine the results of principal curvature estimation using the synthetic datasets shown in Figure \ref{eq:pose_update}. As described in Section \ref{sec:testing_methodolgy} we compare both of our novel joint approaches to existing methods, including a quadric least-squares (\emph{Quad LS}) approach \cite{Douros2002}, and our previous iterative quadric fitting (\emph{Quad IT}) approach from \cite{Spek2015}. We show the results of testing on synthetic data in Figure \ref{fig:curv_err_synth}. Not so surprisingly using information from multiple frames results in reduced error, with both \emph{J-full} and \emph{J-ftf} consistently out-performing the previous methods. Our \emph{J-ftf} method only uses two frames and as such the improvements are considerably more modest then the multi-frame J-full. 

\subsubsection{Real-World Dataset Evaluation}
Finally we compare the implementation on real-world datasets generated for \cite{Spek2015}, an example is shown in Figure \ref{fig:curv_dataset}. The result of this testing shown in Figure \ref{fig:curv_err_real} demonstrates both our joint methods (\emph{J-ftf}, \emph{J-full}) can out-perform previous methods at principal curvature estimation on real-world data. This is significant as curvature can be used as a basis for scene segmentation, using our \emph{J-ftf} method can therefore greatly improve the results of segmentation. We also include a qualitative example of the improvement in Figure \ref{fig:qualitative}, which clearly demonstrates the improvement using multiple frames has. This frame is from the dataset \emph{Real-World 4}, and contrasts the performance of our previous approach (\emph{Quad IT}) against our full joint optimisation (\emph{J-full}). A more subtle aspect of the improvement is the reduction in noise on the planar surface, which is much flatter for the joint estimate. 


\section{Conclusions and Future Work}

We present two systems that use a joint optimisation approach to improve curvature and pose estimates on synthetic and real-world datasets. We show that this approach can not only be used to improve offline dataset accuracy over state of the art techniques such as dense ICP bundle adjustment but can also be incorporated into a real-time system to greatly reduce pose drift to improve the results of re-localisation and loop-closure. We provide all described systems as open-source to be used freely in robotic vision applications, which will be provided upon publication. The current implementation of joint optimisation across multiple frames has not been fully optimised, but has been designed such that a GPU implementation should be a relatively simple extension of this work.
%
%

\section{Acknowledgement}
This work was supported by the Australian Research Council Centre of Excellence for Robotic Vision (project number CE140100016).


%
%



\begin{thebibliography}{10}
	
	\small
	
	\providecommand{\url}[1]{#1}
	\csname url@rmstyle\endcsname
	\providecommand{\newblock}{\relax}
	\providecommand{\bibinfo}[2]{#2}
	\providecommand\BIBentrySTDinterwordspacing{\spaceskip=0pt\relax}
	\providecommand\BIBentryALTinterwordstretchfactor{4}
	\providecommand\BIBentryALTinterwordspacing{\spaceskip=\fontdimen2\font plus
		\BIBentryALTinterwordstretchfactor\fontdimen3\font minus
		\fontdimen4\font\relax}
	\providecommand\BIBforeignlanguage[2]{{%
			\expandafter\ifx\csname l@#1\endcsname\relax
			\typeout{** WARNING: IEEEtran.bst: No hyphenation pattern has been}%
			\typeout{** loaded for the language `#1'. Using the pattern for}%
			\typeout{** the default language instead.}%
			\else
			\language=\csname l@#1\endcsname
			\fi
			#2}}
	
	\bibitem{Guillaume2004}
	L.~Guillaume, ``{Curvature Tensor Based Triangle Mesh Segmentation with
		Boundary Rectification},'' in \emph{Proceedings Computer Graphics
		International (CGI)}, 2004.
	
	\bibitem{Rusinkiewicza}
	S.~Rusinkiewicz, ``{Estimating curvatures and their derivatives on triangle
		meshes},'' in \emph{Symposium on 3D Data Processing, Visualization and
		Transmission (3DPVT)}, no.~2.\hskip 1em plus 0.5em minus 0.4em\relax IEEE,
	2010, pp. 486--493.
	
	\bibitem{Mitra2004}
	N.~J. Mitra, N.~Gelfand, H.~Pottmann, and L.~Guibas, ``{Registration of point
		cloud data from a geometric optimization perspective},'' in
	\emph{Eurographics/ACM SIGGRAPH symposium on Geometry processing}, no.
	January, 2004, p.~22.
	
	\bibitem{Rusu2010a}
	R.~B. Rusu, ``{Semantic 3D Object Maps for Everyday Manipulation in Human
		Living Environments},'' Ph.D. dissertation, 2010.
	
	\bibitem{Sanchez-Fibla2013}
	M.~Sanchez-Fibla, A.~Duff, and P.~F. Verschure, ``{A sensorimotor account of
		visual and tactile integration for object categorization and grasping},'' in
	\emph{IEEE International Conference on Robotics and Automation (ICRA)}.\hskip
	1em plus 0.5em minus 0.4em\relax IEEE, May 2013, pp. 107--112.
	
	\bibitem{Lee2014}
	J.~Lee, S.~Kim, and S.-J. Kim, ``{Mesh segmentation based on curvatures using
		the GPU},'' \emph{Multimedia Tools and Applications}, no. April, Jun 2014.
	
	\bibitem{Khoshelham2012}
	K.~Khoshelham and S.~O. Elberink, ``{Accuracy and resolution of Kinect depth
		data for indoor mapping applications.}'' \emph{Sensors}, vol.~12, no.~2, pp.
	1437--54, Jan 2012.
	
	\bibitem{Spek2015}
	A.~Spek and T.~Drummond, ``{A Fast Method For Computing Principal Curvatures
		From Range Images},'' in \emph{Australasian Conference on Robotics and
		Automation (ACRA)}.\hskip 1em plus 0.5em minus 0.4em\relax ARAA, 2015.
	
	\bibitem{Besl1988}
	P.~Besl and R.~Jain, ``{Segmentation through variable-order surface fitting},''
	\emph{IEEE Transactions on Pattern Analysis and Machine Intelligence (PAMI)},
	vol.~10, no.~2, pp. 167--192, mar 1988.
	
	\bibitem{Douros2002}
	I.~Douros and B.~Buxton, ``{Three-Dimensional Surface Curvature Estimation
		using Quadric Surface Patches},'' \emph{Scanning}, vol.~44, no.~0, 2002.
	
	\bibitem{Griffin2012}
	W.~Griffin, Y.~Wang, D.~Berrios, and M.~Olano, ``{Real-time GPU surface
		curvature estimation on deforming meshes and volumetric data sets.}''
	\emph{IEEE Transactions on Visualization and Computer Graphics}, vol.~18,
	no.~10, pp. 1603--13, Oct 2012.
	
	\bibitem{Besl1992}
	P.~J. Besl and N.~D. McKay, ``{A Method for Registration of 3-D Shapes},''
	\emph{IEEE Transactions on Pattern Analysis and Machine Intelligence (PAMI)},
	vol.~14, no.~2, pp. 239--256, 1992.
	
	\bibitem{Chen1991}
	Y.~Chen and G.~Medioni, ``{Object Modeling by Registration of Multiple Range
		Images},'' in \emph{IEEE International Conference on Robotics and Automation
		(ICRA)}, 1991.
	
	\bibitem{Newcombe2011}
	R.~a. Newcombe, A.~J. Davison, S.~Izadi, P.~Kohli, O.~Hilliges, J.~Shotton,
	D.~Molyneaux, S.~Hodges, D.~Kim, and A.~Fitzgibbon, ``{KinectFusion:
		Real-time dense surface mapping and tracking},'' in \emph{IEEE International
		Symposium on Mixed and Augmented Reality (ISMAR)}.\hskip 1em plus 0.5em minus
	0.4em\relax IEEE, Oct 2011, pp. 127--136.
	
	\bibitem{Salas-Moreno2013}
	R.~F. Salas-Moreno, R.~A. Newcombe, H.~Strasdat, P.~H.~J. Kelly, and A.~J.
	Davison, ``{SLAM++: Simultaneous localisation and mapping at the level of
		objects},'' in \emph{IEEE Conference on Computer Vision and Pattern
		Recognition (CVPR)}.\hskip 1em plus 0.5em minus 0.4em\relax IEEE, 2013, pp.
	1352--1359.
	
	\bibitem{Rusinkiewicz}
	S.~Rusinkiewicz and M.~Levoy, ``{Efficient variants of the ICP algorithm},'' in
	\emph{IEEE International Conference on 3-D Digital Imaging and Modeling
		(3DIM)}.\hskip 1em plus 0.5em minus 0.4em\relax IEEE Comput. Soc, 2001, pp.
	145--152.
	
	\bibitem{Newcombe2011b}
	R.~a. Newcombe, S.~J. Lovegrove, and A.~J. Davison, ``{DTAM: Dense tracking and
		mapping in real-time},'' in \emph{IEEE International Conference on Computer
		Vision (ICCV)}.\hskip 1em plus 0.5em minus 0.4em\relax IEEE, Nov 2011, pp.
	2320--2327.
	
	\bibitem{Whelan12rssw}
	T.~Whelan, M.~Kaess, M.~Fallon, H.~Johannsson, J.~Leonard, and J.~McDonald,
	``Kintinuous: Spatially extended {K}inect{F}usion,'' in \emph{RSS Workshop on
		RGB-D: Advanced Reasoning with Depth Cameras}, Sydney, Australia, Jul 2012.
	
	\bibitem{Izadi2011}
	S.~Izadi, R.~a. Newcombe, D.~Kim, O.~Hilliges, D.~Molyneaux, S.~Hodges,
	P.~Kohli, J.~Shotton, A.~J. Davison, and A.~Fitzgibbon, ``{KinectFusion:
		real-time dynamic 3D surface reconstruction and interaction},'' in \emph{ACM
		SIGGRAPH 2011 Talks}, 2011, p.~23.
	
	\bibitem{Guillaume2004a}
	L.~Guillaume, D.~Florent, and B.~Atilla, ``{Curvature Tensor Based Triangle
		Mesh Segmentation with Boundary Rectification},'' in \emph{Proceedings
		Computer Graphics International (CGI)}.\hskip 1em plus 0.5em minus
	0.4em\relax IEEE, 2004, pp. 10--25.
	
	\bibitem{Griffin2011}
	W.~Griffin, Y.~Wang, D.~Berrios, and M.~Olano, ``{GPU Curvature Estimation on
		Deformable Meshes},'' in \emph{Symposium on Interactive 3D Graphics and
		Games}, 2011, pp. 159--166.
	
	\bibitem{helgason1979differential}
	S.~Helgason, \emph{Differential geometry, Lie groups, and symmetric
		spaces}.\hskip 1em plus 0.5em minus 0.4em\relax Academic press, 1979,
	vol.~80.
	
	\bibitem{m_est}
	\BIBentryALTinterwordspacing
	Z.~Zhang. (1996) M-estimators. [Online]. Available:
	\url{http://research.microsoft.com/en-us/um/people/zhang/INRIA/Publis/Tutorial-Estim/node24.html}
	\BIBentrySTDinterwordspacing
	
	\bibitem{Ouellette1981}
	D.~V. Ouellette, ``{Schur complements and statistics},'' \emph{Linear Algebra
		and Its Applications}, vol.~36, no.~C, pp. 187--295, 1981.
	
	\bibitem{Triggs2000}
	B.~Triggs, P.~Mclauchlan, R.~Hartley, and A.~Fitzgibbon, ``{Bundle Adjustment
		— A Modern Synthesis},'' \emph{Vision Algorithms: Theory {\&} Practice},
	vol. 34099, pp. 1--71, 2000.
	
\end{thebibliography}
\end{document}